\documentclass[10pt,twocolumn,letterpaper]{article}

\usepackage{cvpr}              % To produce the CAMERA-READY version
\usepackage{color}

\usepackage{multirow} 
\usepackage{colortbl}  %彩色表格需要加载的宏包
\usepackage{xcolor}
\definecolor{LemonChiffon}{rgb}{1,0.98,0.8}
\definecolor{YourPink}{rgb}{1,0.8,0.8}

\definecolor{cvprblue}{rgb}{0.21,0.49,0.74}
\usepackage[pagebackref,breaklinks,colorlinks,allcolors=cvprblue]{hyperref}

\def\paperID{15020} % *** Enter the Paper ID here
\def\confName{CVPR}
\def\confYear{2025}

\title{MetricGrids:  Arbitrary Nonlinear Approximation with Elementary Metric Grids based Implicit Neural Representation}

\author{Shu Wang$^{1}$\footnotemark[1]
\qquad
Yanbo Gao$^{1}$\thanks{Equal contributions.}
\qquad
Shuai Li$^{1}$\thanks{Corresponding author.}
\qquad
Chong Lv$^{1}$\qquad
Xun Cai$^{1}$\\
Chuankun Li$^{2}$
\qquad
Hui Yuan$^{1}$
\qquad
Jinglin Zhang$^{1}$\\
$^{1}$Shandong University \qquad $^{2}$North University of China\\
}

\begin{document}
\maketitle
\begin{abstract}

This paper presents MetricGrids, a novel grid-based neural representation that combines elementary metric grids in various metric spaces to approximate complex nonlinear signals. While grid-based representations are widely adopted for their efficiency and scalability, the existing feature grids with linear indexing for continuous-space points can only provide degenerate linear latent space representations, and such representations cannot be adequately compensated to represent complex nonlinear signals by the following compact decoder. To address this problem while keeping the simplicity of a regular grid structure, our approach builds upon the standard grid-based paradigm by constructing multiple elementary metric grids as high-order terms to approximate complex nonlinearities, following the Taylor expansion principle. Furthermore, we enhance model compactness with hash encoding based on different sparsities of the grids to prevent detrimental hash collisions, and a high-order extrapolation decoder to reduce explicit grid storage requirements. experimental results on both 2D and 3D reconstructions demonstrate the superior fitting and rendering accuracy of the proposed method across diverse signal types, validating its robustness and generalizability. Code is available at \href{https://github.com/wangshu31/MetricGrids}{https://github.com/wangshu31/MetricGrids}.
\end{abstract}
\section{Introduction}
% \label{sec:intro}

 \begin{figure}[t]
  \centering
  % \fbox{\rule{0pt}{2in} \rule{0.9\linewidth}{0pt}}
\includegraphics[width=0.98\linewidth]{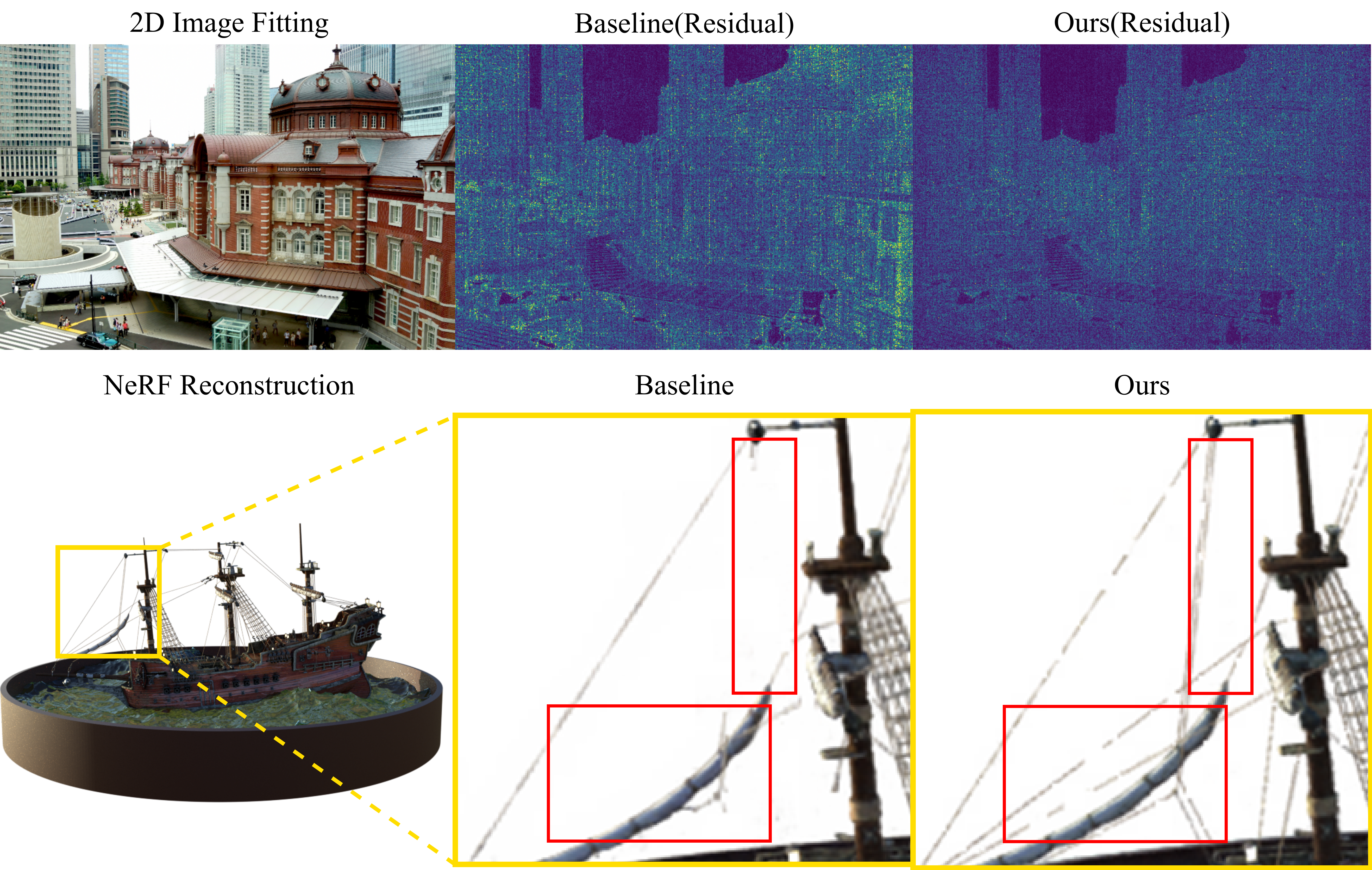}
    \vspace{-2mm}
   \caption{Illustration of the nonlinear fitting capabilities for various type of signals using the existing INR method and our MetricGrids.}
   \label{fig:start}
   \vspace{-7mm}
\end{figure}

Implicit neural representations (INRs)~\cite{sitzmann2020implicit} have emerged as a novel paradigm for signal representation, demonstrating superior performance in reconstructing various signals, including 2D images, 3D videos, 3D shapes, and radiance fields~\cite{chen2021learning,chen2021nerv,sitzmann2019scene,chen2019learning,mildenhall2021nerf}. INRs leverage neural networks to establish a mapping between continuous coordinates and target outputs by employing regularly sampled signal values as training data and optimizing the network parameters. The original INRs utilize a fully connected network with parameters shared across the entire input domain, resulting in a compact model~\cite{chabra2020deep,park2019deepsdf}. However, its expressive ability is constrained by spectral bias~\cite{rahaman2019spectral,tancik2020fourier,bietti2019inductive} and the capacity of the network~\cite{hornik1991approximation}, making it challenging to fit high-frequency and complex signals. To address this limitation, hybrid representations have been proposed~\cite{sitzmann2019deepvoxels,takikawa2021neural,sun2022direct}, which store latent space features at discrete vertex coordinates and use a decoder for reconstruction. Features are typically stored in grid-based structures, such as feature maps for 2D signals and feature volumes for 3D signals~\cite{jiang2020local,liu2020neural}. The input coordinates can be used to directly retrieve the corresponding features through indexing rules. 

In order to represent complex details, hybrid representations need to establish a dense feature grid in the entire signal input domain to ensure that the feature at continuous-space points can be obtained with high accuracy based on the stored points. Especially considering that different signals show different complexity and sparsity among the space, a regular grid-based representation often leads to large storage and redundancy. Various methods have been proposed for efficient grid structure. Early methods focused on modifying grid structures through heuristic modifications such as pruning and merging to create signal-specific data structures~\cite{takikawa2021neural,martel2021acorn,yu2021plenoctrees}. More recent advancements have simplified grid structures to reduce storage demands~\cite{chen2022tensorf,chen2023factor,chibane2020implicit}. There are also methods combining the grid with different representations, such as NeuRBF~\cite{chen2023neurbf} using both adaptive radial bases and grids. However, these methods still require dense features in the complex signal regions, and enhancing the representation accuracy of grid-based methods remains challenging. Especially how to effectively approximate a continuous-space nonlinear signal with a discrete grid representation has not been thoroughly investigated yet.

This paper first investigates the mechanism of using grid structure for signal approximation, and shows that the regular grid relying on linear interpolation implicitly constrains the latent feature space to a linear form, which is a degenerate case of the nonlinear signal. Consequently, for complex nonlinear signals, the features at continuous-space points that are not stored in the grid cannot be accurately represented with the grid, leading to representation errors as shown in \cref{fig:start}.  To solve this problem, we propose MetricGrids, which leverages the concept of Taylor expansion by constructing multiple elementary metric grids in various metric spaces. Each elementary metric grid serves as a high-order approximation of the latent feature space. Subsequently, a compact representation of the elementary metric grids is developed to explore the sparsity of high-order derivatives, thereby maintaining the model compactness. Furthermore, to reduce the number of required grids while achieving complex nonlinear transformations in the latent space, a high-order extrapolation decoder is used to progressively generate higher-order terms based on the learned metric grids. 

The contributions of this paper can be summarized as follows. \begin{itemize}
\item We propose a novel method, MetricGrids, to enhance the grid-based neural representations. It approximates complex nonlinearities of latent feature spaces using elementary metric grids in the form of Taylor expansion.
\item A hash encoding based compact representation of the elementary metric grids is provided to improve parameter efficiency, which explores the sparsity of the high-order derivatives. 
\item We propose a high-order extrapolation decoder to predict the high-order terms based on the low-order approximation obtained from the learned elementary metric grids. 
\item Extensive experiments on multiple fundamental INR tasks have been conducted and the proposed MetricGrids achieves state-of-the-art performance while keeping the model compactness. 
\end{itemize}
\section{Related Work}
\label{sec:relate}

 \textbf{Implicit Neural Representations.} Implicit neural representations (also called neural fields) use neural networks to represent signals such as the 3D shape~\cite{chen2019learning,park2019deepsdf,Mescheder_2019_CVPR}. INRs typically take 3D coordinates as input and produce signed distance function (SDF) or occupancy values.  In~\cite{sitzmann2020implicit}, Sitzmann \etal introduced SIREN, extending implicit neural representations to 2D image representation. Later Mildenhall \etal proposed Neural Radiance Fields (NeRF)~\cite{mildenhall2021nerf} to render a complete 3D scene with INR and volumetric rendering. As the complexity of represented signals increases, many methods have been proposed to improve the representation accuracy, including using different activation functions ~\cite{ramasinghe2022beyond,saragadam2023wire,liu2024finer}, altering the positional encoding basis function~\cite{tancik2020fourier,wang2021spline}, or progressively increasing the positional encoding frequency~\cite{hertz2021sape}. Different neural network architectures have also been developed. In ~\cite{fathony2020multiplicative}, Multiplicative Filter Networks (MFN) was used to represent the reconstructed signal by multiplying together sinusoidal or Gabor wavelet functions applied to the input. Based on this, BACON~\cite{lindell2022bacon}, PNF~\cite{yang2022polynomial}, and RMFN~\cite{shekarforoush2022residual} were proposed by carefully controlling the bandwidth of each filter layer to achieve multi-scale signal reconstruction. Fourier reparameterization~\cite{shi2024improved} and batch normalization ~\cite{cai2024batch} were further proposed to improve the weight composition and learning process, in order to mitigate the spectral bias of networks. A comprehensive explanation of the fully implicit representation approach from the perspective of structured signal dictionaries was presented in~\cite{yuce2022structured}, establishing a connection between network parameters and frequency support. 

\textbf{Hybrid Representations.} Another approach to address spectral bias is to localize certain network parameters by utilizing a feature grid to capture and store local features of specific signal coordinates. Since these features are directly related to the spatial location of the signal, this approach is also referred to as hybrid or grid-based representation ~\cite{jiang2020local,peng2020convolutional,sun2022direct}. Such methods acquire features of any continuous point through indexing the grid with linear interpolation and use them as the latent vector for MLP decoder. However, its expressiveness is closely tied to grid size, often resulting in high memory consumption. Various techniques have been developed to reduce the grid size, including multi-resolution orthogonal planes~\cite{chan2022efficient,fridovich2023k,shue20233d,zou2024triplane}, sparse octree structures~\cite{yang2023tinc,yu2021plenoctrees}, hash encoding~\cite{muller2022instant}, tensor decomposition~\cite{chen2022tensorf} and wavelet transform~\cite{rho2023masked}, etc. Despite these advancements, most methods only focus on reducing grid size and lack the study on improving the fitting accuracy.

 \begin{figure*}[t]
  \centering
  % \fbox{\rule{0pt}{2in} \rule{0.9\linewidth}{0pt}}
   \includegraphics[width=0.88\linewidth]{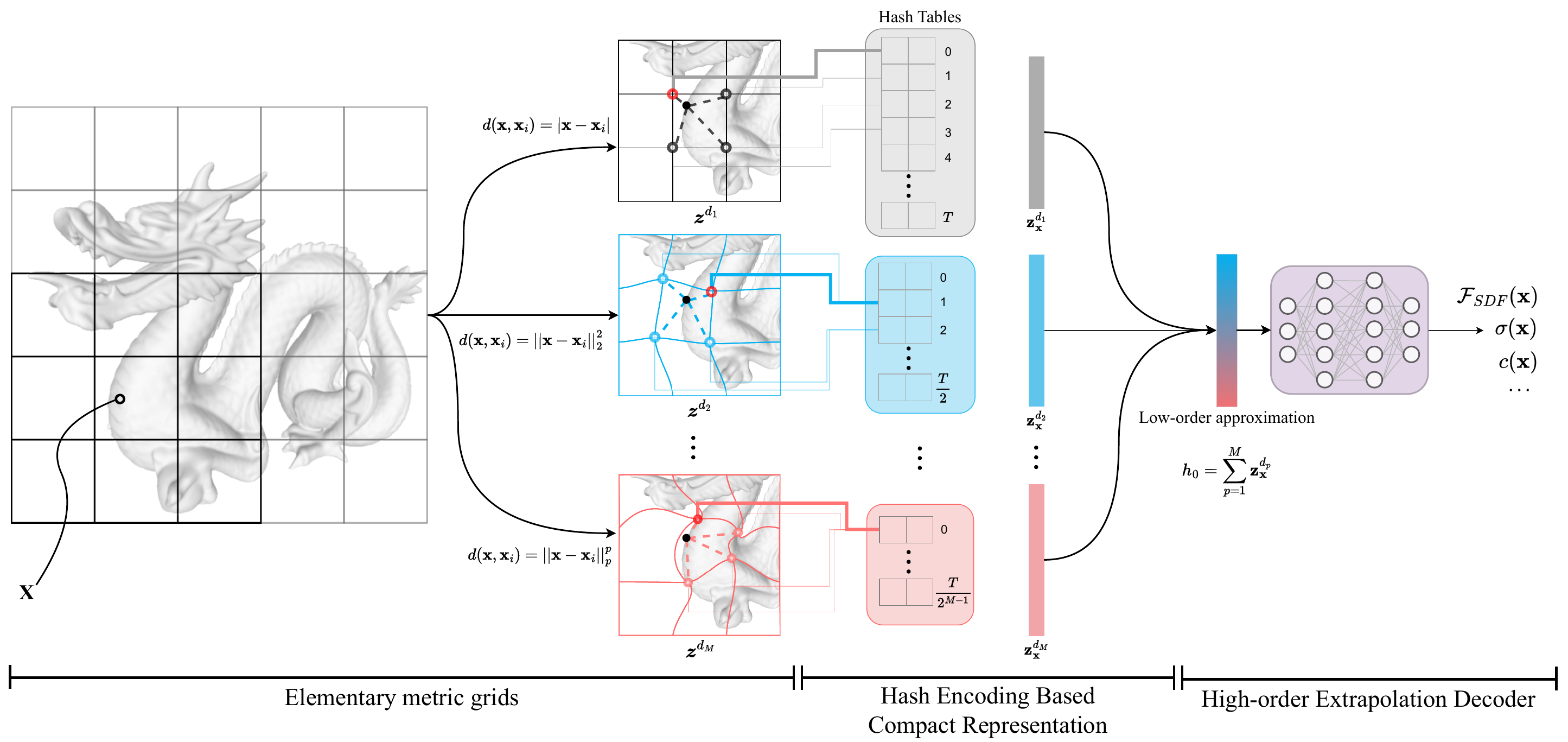}
   \vspace{-2mm}
   \caption{Illustration of the proposed \textbf{MetricGrids}. Given an input coordinate x, features are indexed in multiple metric grids, representing the different-order approximations as   Taylor expansion. Hash encoding is used based on the sparsity of each metric grid to form a compact representation.   A high-order extrapolation decoder is used to further estimate the remaining high-order terms not stored in the metric grids, thereby enhancing the nonlinear fitting capability.
}
   \label{fig:Illustration}
   \vspace{-5mm}
\end{figure*}

There are also methods investigating using different types of local representations, instead of a regular grid. Some methods employ MLPs to learn local structures. KiloNeRF~\cite{reiser2021kilonerf} proposed to directly replace the feature vectors stored in the grid with local MLPs. In~\cite{saragadam2022miner} and~\cite{hao2022implicit}, a pyramid structure and levels of experts were proposed to represent different scales of the local features, respectively.  Different from using MLPs, 3DGS~\cite{kerbl20233d} employed explicitly learned Gaussian primitives at different locations for representation, which has emerged as a pioneering paradigm in the field of radiance fields and achieves remarkable outcomes. NeuRBF~\cite{chen2023neurbf} also proposed to use adaptive radial basis function based representation to enhance the grid. On the other hand, there are also methods of relaxing the grid structures by utilizing point clouds~\cite{li2022learning,xu2022point,zhang2024papr,lu2024scaffold} to store scene features.  In contrast to the above methods, this paper aims to improve the standard grid-based paradigm in terms of fitting accuracy, and develops a novel MetricGrids framework inspired by the Taylor expansion to improve the nonlinear approximation accuracy. 

\section{Proposed Method}
\label{sec:method}

Our goal is to represent a continuous signal with a discrete grid-based neural representation. Naturally, it is an approximation problem and in this paper, we propose MetricGrids, by combining a series of elementary metric grids to solve this problem.

\subsection{Formulation of MetricGrids}
 Given a signal $\mathbf{y}$  with a set of observations, an implicit neural representation is defined as a mapping from input to output, parameterized using a neural network: $\hat{\mathbf{y}}=f_{\Theta}(\mathbf{x}){:}\mathbb{R}^{d}\to\mathbb{R}^{out}$, where $\mathbf{x}\in\mathbb{R}^d$ and $\hat{\mathbf{y}}\in\mathbb{R}^{out}$ denote the input (typically coordinates) and reconstructed output, respectively, $\Theta$ denotes the learnable parameters. The objective is to reconstruct the signal, $i.e.$, minimize the distance $|\left|\mathbf y-\hat{\mathbf{y}}\right||$~\cite{xie2023diner,maiya2024latent}. For fully implicit representation, a neural network like MLP is used to map the coordinates directly to reconstruct the signal, usually exhibiting a strong spectral biass~\cite{rahaman2019spectral,tancik2020fourier,bietti2019inductive} towards the low frequency.
 
 The grid-based representation, also known as the hybrid representation, mitigates this problem by localizing most of the parameters as a feature grid. It can be expressed as a two-stage model:
\begin{equation}
\begin{aligned}
\mathbf{z_{x}}=g(\mathbf{x},\boldsymbol{\mathcal{Z}}), \\
 \hat{\mathbf{y}}=f_{d}(\mathbf{z_{x}},\theta_{d}),
\end{aligned}
\end{equation}
where learnable parameters contain a feature grid and a decoder $\Theta=\{\boldsymbol{\mathcal{Z}},\theta_{d}\}$. The grid $\boldsymbol{\mathcal{Z}}\in\mathbb{R}^{N\times C_{h}}$ is defined on the discrete coordinates of the input domain consisting of learned features on the vertex. $g(\cdot)$ is a distance-dependent (typically $L1$ distance $|\mathbf{x}-\mathbf{x}_i|$) indexing function to interpolate the feature at any point $\mathbf{x}$ in the continuous signal. The decoder $f_{d}$ typically utilizes a compact MLP, similar to fully implicit methods. The grid employs an explicit geometric data structure to store local information of the signal, thereby avoiding the spectral bias problem. It also accelerates the inference, since the indexing function usually does not require network inference~\cite{sun2022direct,dou2023multiplicative,wu2023neural}. 

From the perspective of signal representation~\cite{van2017neural}, the grid can be considered as a discrete feature representation of a continuous signal. The indexing process is to generate the feature at any continuous location based on the discrete grid, and the linear interpolation based generation indicates a linear approximation of the signal. However,  natural scenes are usually complex and highly nonlinear, and thus correspond to nonlinear feature spaces. The conventional grid based paradigm only employs a feature grid constructed in a degenerate space that is a linear approximation of the scene. In regions with complex signals, this degenerate representation presents challenges, and cannot achieve good performance as shown in \cref{fig:start}. 

To solve the above problem, the proposed MetricGrids uses a series of elementary metric grids to generate complex nonlinear representations. In view that any point of a continuous function can be approximated at its neighboring point with a polynomial containing different order terms using Taylor expansion, any point in a grid can also be approximated by a series of elementary metric grids. Each elementary metric grid contains the derivative-related information as the different order terms in Taylor expansion and the distance between points are calculated with metrics of different orders. By learning such an elementary metric grid, it can provide a form of $\frac{z^{(n)}(x_i)}{n!}(x-x_i)^n$, where $x$ and $x_i$ represent a point and its neighboring vertex point stored in a grid, respectively, $z^{(n)}(x_i)$ represent the $n$-th order derivate at the vertex point $x_i$ and $(x-x_i)^n$ can be conveniently generated by a $n$-th order metric defined on the grid. For high-dimensional space, it can be similarly deduced by approximating the sum of polynomials in each direction with the defined metric $\sum_{j}{\frac{z^{(n)}(x_i)}{n!}(x_j-{x_i}_j)^n\approx C_n\mid\mid\mathbf{x}-\mathbf{x}_i\mid\mid^n}$ since the $n$-th order metric accounts for the distances of varying orders in all dimensions $j$.

By summarizing the elementary metric grids, the proposed MetricGrids can provide accurate features for continuous-space that are not directly stored in the grid. In order to reduce the storage of multiple grids, a hash encoding based compact representation is used to fully exploit the sparsity of high-order terms. Moreover,  to further reduce the number of required elementary metric grids, the high-order grids are no longer explicitly stored and instead a high-order extrapolation decoder is designed to approximate them based on the stored lower-order grids. 

In the following, \cref{sec:elementary} introduces the elementary metric-defined derivative grids, \cref{sec:hash_encode} describes the hash encoding based compact representation, and \cref{sec:extrapolation} explains the high-order extrapolation based decoding. \cref{fig:Illustration} illustrates the proposed framework.

\subsection{Elementary Metric Grids }\label{sec:elementary}
 To obtain a nonlinear approximation of a point in the scene, multiple elementary grids are used in different metric spaces, resulting in a series of elementary metric grids. Specifically, the distances in various nonlinear metric spaces are used as proxies of the polynomial basis $(\mathbf{x}-\mathbf{x}_i)^n$, while the grids are learned to store different order derivatives related features.  

Accordingly, a series of elementary metric grids are constructed with different metrics as 
\begin{equation}
\begin{aligned}
\boldsymbol{\mathcal{Z}}^{metrics}=[\boldsymbol{\mathcal{Z}}^{d_{1}},\boldsymbol{\mathcal{Z}}^{d_{2}},...,\boldsymbol{\mathcal{Z}}^{d_{M}}]\in\mathbb{R}^{M\times N\times C_{h}}, \\
\textrm{where}\  \boldsymbol{\mathcal{Z}}^{d_p},d_p({\mathbf{x}},\mathbf{x}_i)=\left|\left|\mathbf{x}-\mathbf{x}_i\right|\right|_p^p,\forall\mathbf{x},\mathbf{x}_i\in\mathbb{R}^d,
\end{aligned}
\label{eq:2}
\end{equation}
where $\boldsymbol{\mathcal{Z}}^{d_p}$ represents a grid structure with discrete vertex coordinate, $d_p(\cdot)$ represents the metric defined on the grid. To align with the Taylor expansion, metrics that are the exponential form of the $p$-norm (to the power of $p$) can be used.

For a point $\mathbf{x}$, our method indexes the metric grids and produces a feature based on the vertex features stored in the grids and distance under diverse metrics as:
\begin{equation}
\begin{aligned}
\boldsymbol{g}(\mathbf{x},\boldsymbol{\mathcal{Z}}^{metrics})=[\mathbf{z}_{\mathbf{x}}^{d_{1}},\mathbf{z}_{\mathbf{x}}^{d_{2}},\ldots,\mathbf{z}_{\mathbf{x}}^{d_{M}}],
\\\textrm{where} \ \ \mathbf{z}_{\mathbf{x}}^{d_{p}}=g\left(\mathbf{x},{\boldsymbol{\mathcal{Z}}^{d_{p}}};d_{p}(\mathbf{x},\mathbf{x}_{i})\right).
\end{aligned}
\end{equation}

With the learning-based grid generation, a high order derivative term $\frac{z^{(n)}(x_{i})}{n!}(x-x_{i})^{n}$ in the Taylor expansion can be directly generated with the grids learning the different derivatives induced features. As a special case, the feature $\mathbf{z}_{\mathbf{x}}^{d_{1}}$ extracted from the first elementary metric grid aligns with the existing grid-based model paradigm by providing a linear approximation of the latent space. In contrast, $\mathbf{z}_{\mathbf{x}}^{d_{m}}\ (m\geq2)$ represents high-order derivative induced features to provide a nonlinear approximation. 
 \begin{figure}[t]
  \centering
  % \fbox{\rule{0pt}{2in} \rule{0.9\linewidth}{0pt}}
   \includegraphics[width=0.98\linewidth]{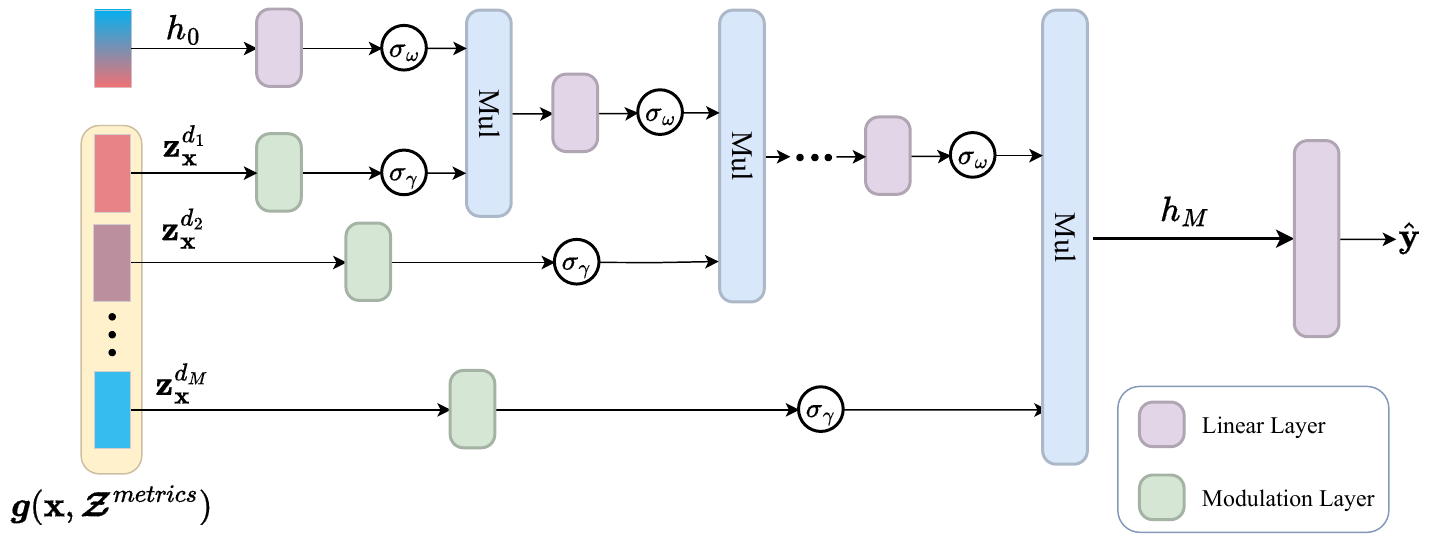}
\vspace{-2mm}
   \caption{Illustration of the high-order extrapolation decoder. The ‘Mul’ represents the Hadamard product.}% , $\sigma_\omega$ and $\sigma_\gamma$ represent the nonlinear activation for backbone and modulation layer, respectively
   \label{fig:decoder}
   \vspace{-5mm}
\end{figure}\

Together, these elementary grids enable latent space features to approximate arbitrary nonlinearities across different locations, thereby facilitating the representation of complex signals. In addition to mimicking the Taylor expansion, different types of metric grids can also be defined as long as the metrics provide a nonlinear distance measure. In such a case, it can also provide a nonlinear approximation of any point within a regularly stored feature grid.

\begin{figure*}[t]
  \centering
  % \fbox{\rule{0pt}{2in} \rule{0.9\linewidth}{0pt}}
   \includegraphics[width=0.98\linewidth]{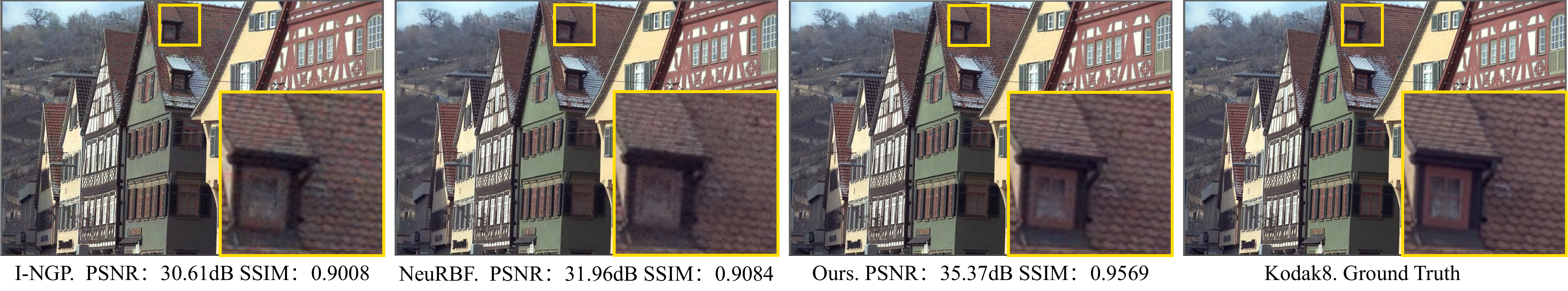}
\vspace{-2mm}
   \caption{Qualitative Comparison on Kodak dataset. Zoomed-in views of windows and roofs illustrate the ability of the baseline model and our proposed MetricGrids in fitting complex details.}
   \label{fig:2d_img}
   \vspace{-5mm}
\end{figure*}

\subsection{Hash Encoding Based Compact Representation }\label{sec:hash_encode}
 In this section, a compact representation of the elementary metric grids is proposed to improve the efficiency of storing and learning multiple grids. Motivated by Instant-NGP~\cite{muller2022instant}, for each vertex coordinate $\mathbf{x}_i$, multiple feature vectors from the elementary metric grids, representing the linear approximation and high-order terms of the latent space need to be stored. Simply taking the elementary metric grids as an extra dimension in addition to the spatial dimensions and creating hash encoding together can cause hash collisions not only in space but also in different elementary metric grids. Especially considering different elementary metric grids containing features of different natures, such collision among elementary metric grids may significantly reduce the representation accuracy. Therefore, the elementary metric grids are encoded separately to avoid collision among grids.

On the other hand, with the elementary metric grids representing the information of linear approximation, and different-order derivatives, the sparsity of the grids is different.  In smooth regions of the signal, the different-order derivatives are all zero, making them much sparser than the linear first-order approximation. Therefore, the elementary metric grids are encoded differently by setting a shorter hash table length T for higher-order metric grids to encourage the fusion of similar features in high-order items. Furthermore, with the nonlinear representation using the elementary metric grids, the high-resolution linear grids are no longer necessary for fitting complex signals. Consequently, the maximum resolution of the grids is also reduced to minimize the number of parameters used in the final representation. For a vertex coordinate $\mathbf{x}_{i}$, the hash mapping at each level can be expressed as:
\begin{equation}
\begin{aligned}
    H(\mathbf{x}_{i}) = [h^{d_{1}}(\mathbf{x}_{i},T),...,h^{d_{M}}(\mathbf{x}_{i},\frac{T}{2^{M-1}})], \\
    \textrm{where}\ h^{d_{m}}(\mathbf{x},T)=\left(\bigoplus_{j=1}^{d}x_{j}\pi_{j}\right)\ \ mod\ \ T, 
\end{aligned}
\end{equation}
where $\oplus$ denotes the bitwise XOR operation and $\pi_i$ are unique, large prime numbers, as in Instant NGP~\cite{muller2022instant}.With the above hash encoding based compression of the elementary metric grids, a compact representation, of the same size as the conventional grid, can be obtained while multiple-order features are extracted to represent each point for better nonlinearity representation.

\begin{figure*}[t]
  \centering
  % \fbox{\rule{0pt}{2in} \rule{0.9\linewidth}{0pt}}
   \includegraphics[width=0.88\linewidth]{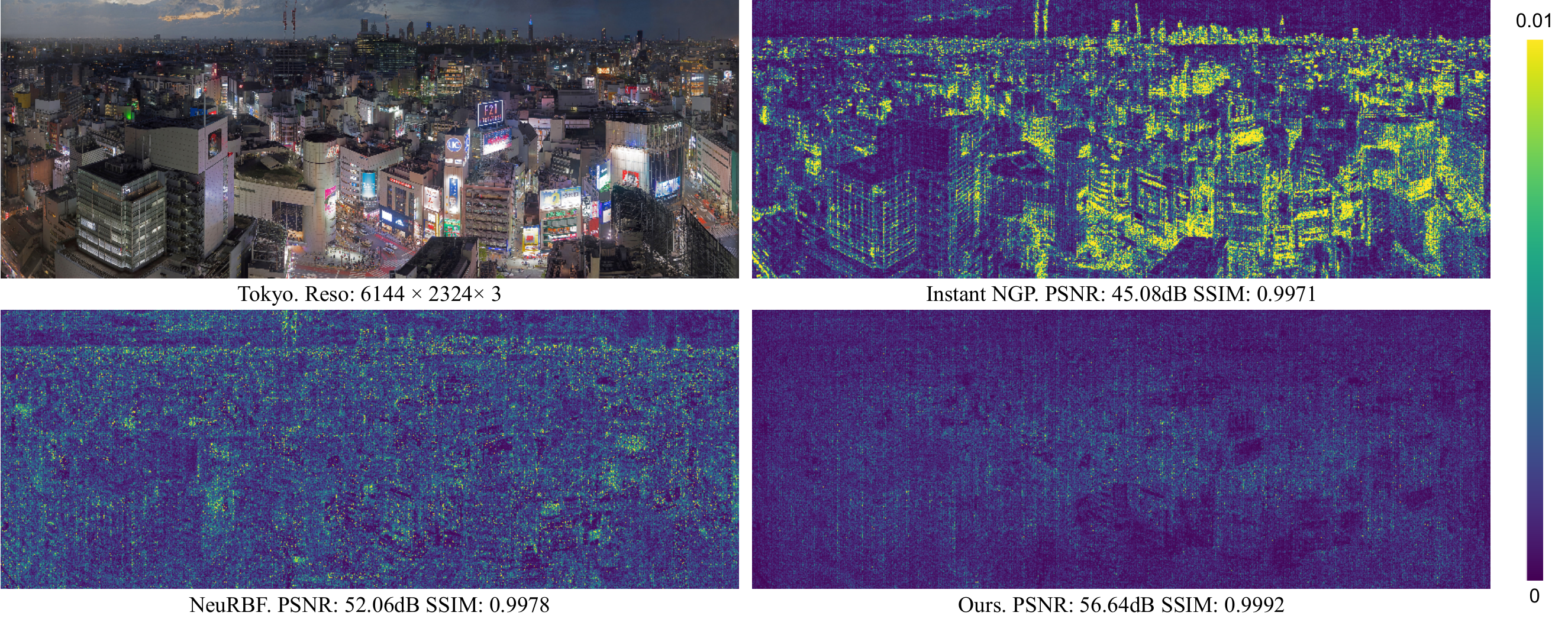}
    \vspace{-2mm}
   \caption{Qualitative Comparison on Gigapixel Image. The upper-right shows the ground truth image, containing numerous intricate details. The upper right, lower-left and lower-right images show L2 error maps for the baselines and our method, respectively. All experiments use consistent hyperparameter settings.}
   \label{fig:gig_fit}
   \vspace{-5mm}
\end{figure*}

\subsection{High-order Extrapolation Decoder }\label{sec:extrapolation}
To reduce the number of learned grids and improve the representation efficiency of the elementary metric grids, only a few grids are explicitly learned in the experiments, three elementary metric grids to be specific. In order to further reduce the approximation error due to the limited number of grids, a high-order extrapolation decoder is developed to generate high-order terms based on the linear and relatively lower-order terms stored in the grids,as shown in \cref{fig:decoder}.

First, the linear and lower-order feature of point $\mathbf{x}$  $\boldsymbol{g}(\mathbf{x},\boldsymbol{\mathcal{Z}}^{metrics})=[\mathbf{z}_{\mathbf{x}}^{d_{1}},\mathbf{z}_{\mathbf{x}}^{d_{2}},\ldots,\mathbf{z}_{\mathbf{x}}^{d_{M-1}}]$ which indexed from the grid are extracted and summarized as the initial approximation of the nonlinear feature. Then a hierarchical extrapolation decoder is developed to estimate the higher-order terms by leveraging the initial approximation and different-order features. Specifically, the features are first modulated with a linear layer and then multiplied with the Hadamard product~\cite{chrysos2021deep,lindell2022bacon,dou2023multiplicative,li2024learning} to generate a higher-order term, and then progressively updated the approximation. For the output features of the decoder layer $\ell$, this process can be expressed as follows:
\begin{equation}
{h}_{\ell+1}=\omega_\ell(h_\ell)\circ\gamma_\ell(\mathbf{x}),
\end{equation}
\begin{equation}
\omega_{\ell}(h_\ell)=\sigma_\omega\left(\boldsymbol{W}_\ell{h}_{\ell-1}+\boldsymbol{b}_\ell\right),
\end{equation}
\begin{equation}
    \gamma_{\ell}(\mathbf{x})=\sigma_\gamma(\boldsymbol{W}_{\ell}^{\gamma}\mathbf{z}_{\mathbf{x}}^{d_{\ell-1}}+\boldsymbol{b}_{\ell}^{\gamma}),
\end{equation}
where $h_\ell\in\mathbb{R}^N$ represents the gradually updated approximation, $\omega_\ell(\cdot)$ and $\gamma_\ell(\cdot)$ are backbone layers and modulation layers with weight $\boldsymbol{W}$ and bias $\boldsymbol{b}$ that adjust the current approximation and higher-order terms. $\sigma_\omega$ and $\sigma_\gamma$ represent nonlinear activation functions, $\sin(\cdot)$ and $\sin^2(\cdot)$ are used in the experiments, as the sine activation function has been demonstrated to outperform common activation functions (e.g. ReLU) in INR tasks~\cite{sitzmann2020implicit}.

Without considering the activation function, the theoretical highest-order term achievable after $M$ decoder layers is $m+1+2+\cdot\cdot\cdot+m=m+\frac{m(m+1)}{2}=\frac{m(m+3)}{2}$, due to the repeated multiplication operations within each layer. Finally, the reconstructed features are converted to the signal domain through a linear layer:
\begin{equation}
\hat{\mathbf{y}}=\boldsymbol{W}_{o}{h}_M+\boldsymbol{b}_{o}.
\end{equation}

This method hierarchically integrates features from each derivative space and extrapolates higher-order terms based on the approximation and lower-order terms, allowing for more complex nonlinear transformations in the latent space.

\section{Experiments}
% \label{sec:Experiment}
\subsection{Implementation Details}
 The proposed MetricGrids framework are evaluated across various Implicit Neural Representation tasks, including 2D image fitting, 3D signed distance function fitting, and 5D neural radiance field reconstruction for novel view synthesis. In the elementary metric grids implementation, three grids are used with the first grid using linear metric as a regular metric while the other two grids with nonlinear metrics. In addition to the nonlinear metrics $[\mathrm{x}^2,\mathrm{x}^3]$, the other nonlinear metrics including $[\sin(\cdot),\ \arcsin(\cdot)]$ and $[\sin(\cdot),\ \cos(\cdot)]$~\cite{mildenhall2021nerf} are also evaluated for 2D image and 3D SDF fitting,  and for neural radiance field reconstruction, respectively. Such nonlinear metrics can actually provide more nonlinear properties in experiments, since they can be further approximated with the polynomial terms. The high-order extrapolation decoder architecture comprised five layers with $64$ hidden units, with the number of layers directly corresponding to the established Elementary Metric Grids.  More details can be found in supplementary.
The training was conducted on a single NVIDIA RTX 3090 GPU using the Adam~\cite{kingma2014adam} optimizer ($\beta_1=0.9, \beta_2=0.99, \epsilon=10^{-15}$) with cosine annealing for learning rate decay. $L2$ loss was employed for 2D image fitting and NeRF reconstruction, while MAPE loss as in ~\cite{muller2022instant} was utilized for 3D SDF reconstruction. The same point sampling strategy as in the Instant NGP ~\cite{muller2022instant} and NeuRBF~\cite{chen2023neurbf} is adopted for SDF fitting experiments.  
\begin{figure*}[t]
  \centering
  % \fbox{\rule{0pt}{2in} \rule{0.9\linewidth}{0pt}}
   \includegraphics[width=0.88\linewidth]{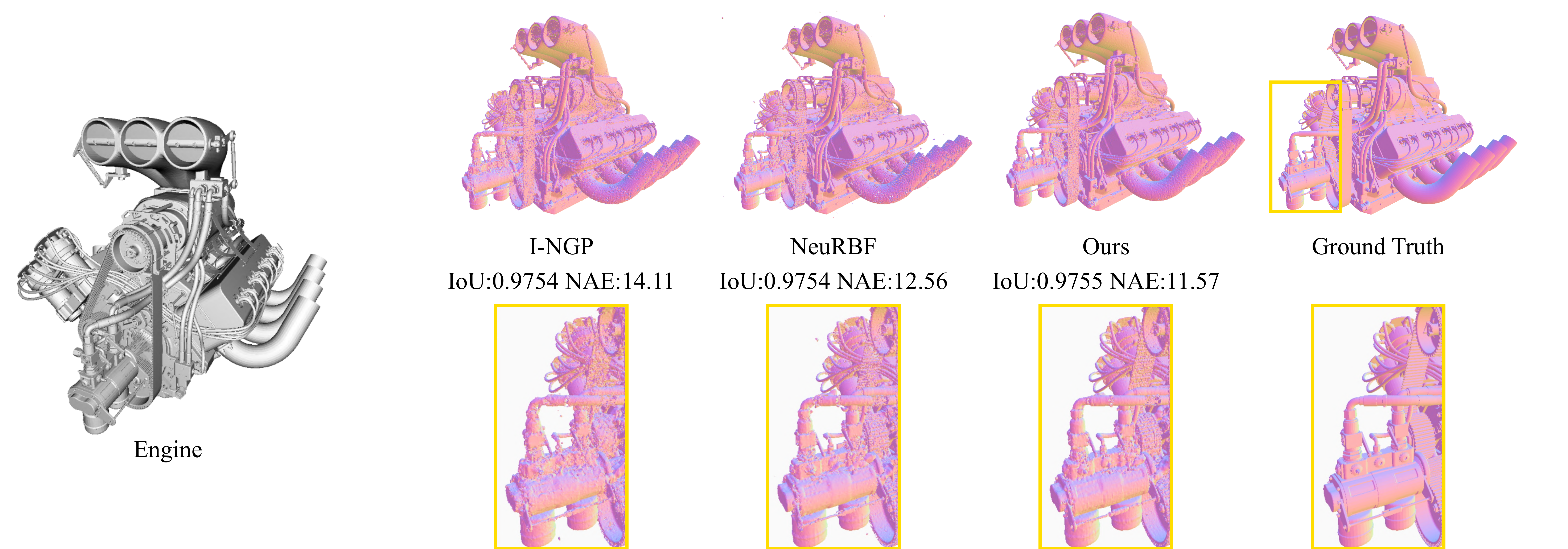}
   \vspace{-2mm}
   \caption{Qualitative Comparison on 3D Signed Distance Field Reconstruction. Details are best viewed with zoom-in.}
   \label{fig:3d_sign}
   \vspace{-3mm}
\end{figure*}

\begin{table}[t]
    \centering
    \renewcommand\arraystretch{1}
    \setlength\tabcolsep{3pt}
    % \resizebox{0.46\textwidth}{!}
{
    \footnotesize
    \begin{tabular}{c|c|ccc}
        \toprule[1pt]
        Method&Rep.& $\textbf{PSNR}\uparrow$& $\textbf{SSIM}\uparrow$& Params(K)   \\
        \midrule
        WIRE~\cite{saragadam2023wire}&  Implicit& 37.59 & 0.9596& 373\\
        SIREN~\cite{sitzmann2020implicit}& Implicit& 38.70 & 0.9512& 207\\
        SCONE~\cite{li2024learning}& Implicit& 40.29 & 0.9690& 207\\
        I-NGP~\cite{muller2022instant}& Hybrid& 37.06& 0.9386& 206 \\
        NFFB~\cite{wu2023neural}& Hybrid& 38.77& 0.9537&  208 \\
        NeuRBF~\cite{chen2023neurbf}& Hybrid& 38.70& 0.9488& 207 \\
        \midrule
        \textbf{Ours}($p$-norm) & Hybrid& \cellcolor{LemonChiffon}40.63& \cellcolor{LemonChiffon}0.9720& 207 \\
        \textbf{Ours} & Hybrid& \cellcolor{YourPink}41.57& \cellcolor{YourPink}0.9735& 207 \\
        \bottomrule[1pt]
    \end{tabular}
    }   
    \vspace{-2mm}
    \caption{Result comparisons of the proposed method against the existing methods under a similar model size, on 2D Image Fitting using Kodak Dataset.\colorbox{YourPink}{Best} and \colorbox{LemonChiffon}{second best} results are highlighted.}
    \label{tab:2d_img}
    \vspace{-5mm}
\end{table}

\subsection{2D Image Fitting}
 We first evaluate the effectiveness of fitting 2D images using the Kodak dataset~\cite{kodak}, which consists of 24 images with a resolution of $768\times512$. PSNR(dB)  and SSIM are used to assess the reconstructed image quality. The performance is first compared with the existing methods with a similar model size (around 207K) and the same training iterations (20,000 steps), the results are shown in \cref{tab:2d_img}. It can be seen that the proposed MetricGrids achieves the best performance compared with other methods. Moreover, it can be seen that while MetricGrids with the $p$-norm based nonlinear metrics$[\mathrm{x}^2,\mathrm{x}^3]$ improves the results, the highly nonlinear metrics $[\sin(\cdot),\ \arcsin(\cdot)]$ perform even better. It is mostly because such nonlinear metrics themselves can be considered as a combination of different-order terms under Taylor expansion, thus help the grid to learn nonlinear approximation. \cref{fig:2d_img} further shows some visual examples. 

Comparison with recent work GaussianImage~\cite{zhang2025gaussianimage} using Gaussian primitives to represent images and other methods with different numbers of parameters are also performed. The results are shown in \cref{tab:GaussianImage}, metrics of the comparison methods are taken from GaussianImage~\cite{zhang2025gaussianimage}. It can be seen that the proposed method with a medium size (m) and large size (l) both significantly improve the performance over GaussianImage~\cite{zhang2025gaussianimage}, especially with the similar number of parameters. 

\begin{table}[t]
    \centering
    \renewcommand\arraystretch{1}
    \setlength\tabcolsep{1pt}
    % \resizebox{0.46\textwidth}{!}
{
    \footnotesize
    \begin{tabular}{c|c|ccc}
        \toprule[1pt]
        Method&Rep.& $\textbf{PSNR}\uparrow$& $\textbf{MS-SSIM}\uparrow$& Params(K)  \\
        \midrule
        I-NGP~\cite{muller2022instant}         & Hybrid   & 43.88 & 0.9976 & 300    \\
        NeuRBF~\cite{chen2023neurbf}        & Hybrid   & 43.78 & 0.9964 & 337    \\
        3D GS~\cite{kerbl20233d} & Explicit & 43.69 & \cellcolor{YourPink}0.9991 & 3540   \\
        GaussianImage~\cite{zhang2025gaussianimage}& Explicit & 44.08 & 0.9985 & 560    \\
        \midrule
        \textbf{Ours-m}& Hybrid   & \cellcolor{LemonChiffon}46.97 & 0.9986 & 335    \\
        \textbf{Ours-l}& Hybrid  & \cellcolor{YourPink}52.22 & \cellcolor{LemonChiffon}0.9987 & 533    \\
        \bottomrule[1pt]
    \end{tabular}
    }   
    \vspace{-2mm}
    \caption{Result comparison with Gaussian primitive based methods, on 2D Image Fitting using Kodak Dataset.}
    \label{tab:GaussianImage}
    \vspace{-2mm}
\end{table}

\begin{table}[t]
    \centering
    \setlength{\tabcolsep}{2mm}
    \renewcommand\arraystretch{1}
    \setlength\tabcolsep{4pt}
    % \resizebox{0.46\textwidth}{!}
{
    \footnotesize
    \begin{tabular}{c|cccc}
        \toprule[1pt]
        Method&$\textbf{IOU} \uparrow$ & $\textbf{NAE}\downarrow$& $\textbf{Chamfer}\downarrow$ & Params    \\
        \midrule
        NGLOD5~\cite{takikawa2021neural}& 0.99991&	4.21&	0.0022& 10.15M \\
        I-NGP~\cite{muller2022instant} & 0.99991&	3.42&	0.0023& 950K \\
        NeuRBF~\cite{chen2023neurbf} & 0.99988&	3.20&	0.0021& 856K \\
        \midrule
        \textbf{Ours} & 0.99991& 3.09& 0.0021& 	905K \\
        \bottomrule[1pt]
    \end{tabular}
    }   
    \vspace{-2mm}
    \caption{Result comparison on 3D Signed Distance Function Reconstruction using Stanford 3D Dataset.}
    \label{tab:3D_SDF}
    \vspace{-6mm}
\end{table}

A notable advantage of the grid-based model over the fully implicit representation is the capacity to accommodate gigapixel images~\cite{martel2021acorn}. To evaluate this scalability, our method is tested on the publicly available gigapixel images, and compared with InstantNGP~\cite{muller2022instant} and NeuRBF~\cite{chen2023neurbf} as representative state-of-the-art baselines, using the same number of parameters. The results are shown in  \cref{fig:gig_fit},  with an illustration of the error map. It can be seen that our method demonstrates superior fidelity in gigapixel image reconstruction compared to InstantNGP~\cite{muller2022instant} and NeuRBF~\cite{chen2023neurbf}. Moreover, the error distribution of the proposed method is relatively uniform across the whole image, indicating that the proposed method can efficiently fit complex regions such as edges with the higher-order grids. Additional results for other gigapixel images are provided in the supplementary material.

\subsection{3D Shape Reconstruction}
 To assess the quality of 3D shape representation, 7 scenes from the Stanford 3D Scanning Repository~\cite{stanford3d} are evaluated, following the hyperparameter configurations established in NeuRBF~\cite{chen2023neurbf}. The evaluation employed multiple metrics: Intersection over Union (IoU), Normal Angular Error (NAE), and Chamfer Distance (CD). The results are demonstrated in \cref{tab:3D_SDF}, and comparative analysis against baseline and state-of-the-art methods reveals that the proposed approach achieves superior performance across all metrics while maintaining comparable model compactness. The qualitative comparisons are presented in \cref{fig:3d_sign}. It can be seen that our method yields a more distinct representation in intricate areas than the other methods.

\begin{figure*}[t]
  \centering
  % \fbox{\rule{0pt}{2in} \rule{0.9\linewidth}{0pt}}
   \includegraphics[width=0.98\linewidth]{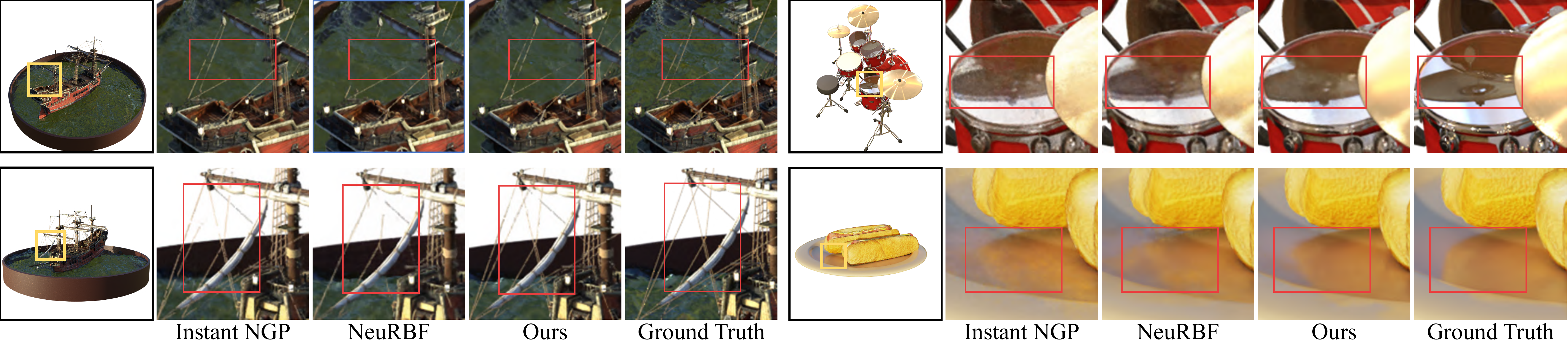}
   \vspace{-2mm}
   \caption{Qualitative Comparison on Neural Radiance Field Reconstruction. Our method consistently outperforms baselines in capturing both complex and fine details in scenes, as well as in smooth regions with reflections.}
   \label{fig:field_recon}
   \vspace{-3mm}
\end{figure*}

\subsection{Neural Radiance Field Reconstruction}
 In the Neural Radiance Field setting, a volumetric shape is represented by a spatial (3D) density function and a spatial-directional (5D) emission function. The Blender dataset~\cite{mildenhall2021nerf} which consists of eight object-centric scenes is widely used as a standard benchmark. We explore using NerfAcc~\cite{li2023nerfacc} as the baseline implementation, while utilizing the same decoder as NebRBF. The experimental settings and hyperparameter selections are outlined in NerfAcc~\cite{li2023nerfacc}. The results are shown in \cref{tab:Synthetic}. It can be seen that under the same number of iterations, our method effectively enhances the rendering quality of new viewpoints while maintaining model compactness. Some visual comparisons are shown in \cref{fig:field_recon}. It can be seen that the proposed method demonstrates consistent improvements in both complex structures and smooth regions.

\begin{table}[t]
    \centering
    \renewcommand\arraystretch{0.9}
    \setlength\tabcolsep{2pt}
    % \resizebox{1\textwidth}{!}
{
    \footnotesize
    \begin{tabular}{c|ccccc}
        \toprule[1pt]
    Methods&Steps& Params&$\textbf{PSNR} \uparrow$ & $\textbf{SSIM}\uparrow$& $\textbf{LPIPS}_\textrm{VGG}  \downarrow$ \\
        \midrule
        Mip-NeRF 360~\cite{barron2022mip}&250k&	3.23M&	33.25&	0.962&	0.039\\
        TensoRF~\cite{chen2022tensorf}& 30k&	17.95M&	33.14&	0.963&	0.047 \\
        K-Planes~\cite{fridovich2023k}& 30k&	33M&	32.36&	0.962&	0.048 \\
        I-NGP~\cite{muller2022instant} & 35k&	12.21M&	33.18&	0.963&	0.051 \\
        NerfAcc~\cite{li2023nerfacc}&	20k&	12.21M&	32.55&	-&	0.056\\
        NeuRBF~\cite{chen2023neurbf} & 30k&	17.74M&	\cellcolor{LemonChiffon}34.62&	\cellcolor{LemonChiffon}0.975&	\cellcolor{LemonChiffon}0.034\\
        \midrule
        $\textbf{Ours}$ & 30k&	13.11M&	 \cellcolor{YourPink}35.02&  \cellcolor{YourPink}0.976& \cellcolor{YourPink}0.033\\
        \bottomrule[1pt]
    \end{tabular}
    }
    \vspace{-2mm}
    \caption{Result comparison on Neural Radiance Field Reconstruction using Blender Dataset~\cite{mildenhall2021nerf}, the symbol “-” indicates that specific information is unavailable in the respective paper.}
    \label{tab:Synthetic}
    \vspace{-6mm}
\end{table}

\begin{table}[t]
    \centering
    \renewcommand\arraystretch{0.9}
    \setlength\tabcolsep{2pt}
    % \resizebox{0.46\textwidth}{!}
{
    \footnotesize
    \begin{tabular}{c|c|cccc|cc}
        \toprule[1pt]
         {\multirow{2}{*} {Settings}} &{\multirow{2}{*} {Decoder}} &\multicolumn{4}{c|}{2D image}&\multicolumn{2}{c}{3D SDF}\\
         % \cmidrule(l{0pt}r{0pt}){3-8}
        &&\textbf{PSNR} & \textbf{SSIM}& Time& Params& 
        \textbf{CD} & 
        \textbf{NAE}\\
        \midrule
       Baseline &MLP& 37.06&	0.9386&	195s&206k&	0.0023&	3.43 \\
        \midrule
        M=2\dag&MLP& 39.45&	0.9580&	336s&234k&	0.0022&	3.34\\
        M=3\dag&MLP& 40.85&	0.9684&	385s&258k&	0.0022&	3.20\\
        M=4\dag&MLP& 41.91&  0.9737&	 576s&279k&	 0.0023&	3.15\\
        \midrule
        M=2&MLP& 38.71&	0.9519&	326s&206k&	0.0022&	3.40\\
        M=3&MLP& 38.95&	0.9538&	428s&206k&	0.0022&	3.35 \\
        M=4&MLP& 38.98&  0.9540&	 534s&206k&	 0.0025&	3.35\\
        \midrule
        Ours&w/o hierarchy&  40.23&	0.9668&	 1136s&  212k& 0.0022 & 3.29\\
        Ours Full&High-order& 41.57&	0.9735&	1148s&207k&	0.0021&	3.09 \\
        \bottomrule[1pt]
    \end{tabular}
    }   
    \vspace{-2mm}
    \caption{Ablation Study. M indicates the number of elementary metric grids, aligned with the definition in \cref{eq:2}. The symbol \dag \ indicates that the overall model size  is not restricted with the increase of grids.} 
    \label{tab:Ablation}
    \vspace{-7mm}
\end{table}

\subsection{Ablation Study}
\label{sec:Ablation}
To validate the effectiveness of each proposed module, ablation studies are conducted on both Kodak dataset~\cite{kodak} for 2D image fitting and Stanford 3D Scanning Repository~\cite{stanford3d} for SDF reconstruction. 

\textbf{Evaluation on elementary metric grids.} The use of elementary metric grids was first evaluated to show the effectiveness of nonlinear approximation. It is built upon the baseline method, using InstantNGP~\cite{muller2022instant}.  We set the highest resolution of the metric grids to half of that used in the I-NGP and reduced the length of the hash table to maintain parameter counts comparable, resulting in metric grids with lower spatial resolution and increased sparsity as described in \cref{sec:hash_encode}. The results are shown in the first section of \cref{tab:Ablation}. We have evaluated the effect of different grid numbers (M) by selecting 2, 3, and 4 metric grids based on the baseline. It can be seen that the elementary metric grids defined in nonlinear metric spaces significantly enhance the performance across all tasks, validating its effectiveness as high-order term approximations. Among the different configurations, M=2 yielded the largest improvement, further underscoring the importance of introducing nonlinear metrics for complex signal fitting.
Additionally, experiments with restricting the total parameters when adding grids are also conducted, as detailed in the central section of \cref{tab:Ablation}. Under the same number of parameters, adding a higher-order term grid effectively improves performance, demonstrating that nonlinear interpolation via metric grids provides a better inductive bias compared to linear interpolation in terms of representation quality. However, performance gains diminish when the number of higher-order terms exceeds three, indicating explicit storage of higher-order terms proves computationally inefficient.  Thus, MetricGrids with M=3 is applied in our implementation.

\textbf{Evaluation on high-order extrapolation decoder.} Based on the previous analysis, we set the order of the elementary metric grid to 3 and incorporate the proposed high-order extrapolation decoder for the implicit estimation of high-order terms at different positions, thereby achieving a balance between fitting accuracy and model compactness, which constitutes our full model. On one hand, it adds more nonlinearity layers to the decoder, improving the decoding accuracy. On the other hand, it generates higher-order terms based on the initial low-order approximation and different grids. To validate this, we masked the hierarchical structure of different grid feature inputs in the decoder (denoted as w/o hierarchy) and demonstrated significantly degraded decoder performance without hierarchical feature integration. In contrast, the high-order extrapolation decoder effectively leverages multiple metric grids to enhance the final performance.

\section{Conclusion}
% \label{sec:conclusion}
 In this paper, we have proposed MetricGrids, which provides accurate neural representations for various signal types. Our analysis reveals that the linear indexing function used to obtain continuous-space point features in the current grid-based neural representation leads to latent space degradation, thus resulting in parameter inefficiency and approximation error on nonlinear signals. Drawing inspiration from Taylor expansion, elementary metric grids defined in nonlinear metric spaces are proposed as high-order terms to capture complex latent space nonlinearity, thereby enhancing representation capabilities. A compact representation is further designed using hash encoding and a high-order extrapolation decoder to improve fitting accuracy without increasing parameter and storage requirements. Experimental results demonstrated that our method significantly outperforms baseline models and achieves state-of-the-art performance in 2D image fitting, 3D shape fitting, and neural radiance field reconstruction tasks.

\noindent\textbf{Acknowledgment.} This work was supported in part by the National Natural Science Foundation of China (No. 62271290 and 62001092).

{
    \small
    \bibliographystyle{ieeenat_fullname}
    \bibliography{main}
}

% WARNING: do not forget to delete the supplementary pages from your submission 
% \input{sec/X_suppl}

\end{document}